\pdfoutput=1

\documentclass[11pt]{article}

\usepackage{acl}

\usepackage{times}
\usepackage{latexsym}

\usepackage[T1]{fontenc}

\usepackage[utf8]{inputenc}

\usepackage{amsmath,amsfonts,amssymb}
\usepackage{times}
\usepackage{latexsym}
\usepackage{tabularx}
\usepackage{pdfpages}
\usepackage{amssymb}
\usepackage{amsmath}
\usepackage{bm}
\usepackage{graphicx}
\usepackage{booktabs}
\usepackage{mathtools}
\usepackage{multirow}
\usepackage{multicol}
\usepackage{makecell}
\usepackage{diagbox}

\title{Mixture-of-LoRAs: An Efficient Multitask Tuning for \\Large Language Models}

\author{Wenfeng Feng\quad
Chuzhan Hao\quad
Yuewei Zhang\quad
Yu Han\quad
Hao Wang\protect \thanks{\ \ Corresponding author}
\\
Alibaba Cloud, Alibaba Group 
\\
\tt {\{wenfeng.fwf,liyou.zyw,hy226261\}@alibaba-inc.com}\\ \tt {chuzhanh@gmail.com,cashenry@126.com}
}

\begin{document}
\maketitle    
\begin{abstract}
Instruction Tuning has the potential to stimulate or enhance specific capabilities of large language models (LLMs). However, achieving the right balance of data is crucial to prevent catastrophic forgetting and interference between tasks. To address these limitations and enhance training flexibility, we propose the \textbf{M}ixture-\textbf{o}f-LoR\textbf{A}s (MoA) architecture – a novel and parameter-efficient tuning method designed for multi-task learning with LLMs.
In this paper, we start by individually training multiple domain-specific LoRA modules using corresponding supervised corpus data. These LoRA modules can be aligned with the expert design principles seen in Mixture-of-Experts (MoE). Subsequently, we combine the LoRAs using an explicit routing strategy and introduce domain labels to facilitate multi-task learning, which helps prevent interference between tasks and ultimately enhances the performance of each individual task. Furthermore, each LoRA model can be iteratively adapted to new domains, allowing for quick domain-specific adaptation. Experiments on diverse tasks demonstrate superior and robust performance of our approach, which will also further promote the application of domain-specific LLMs.
\end{abstract}

\section{Introduction}
\begin{figure}[ht]
\centering
\includegraphics[width=0.48\textwidth]{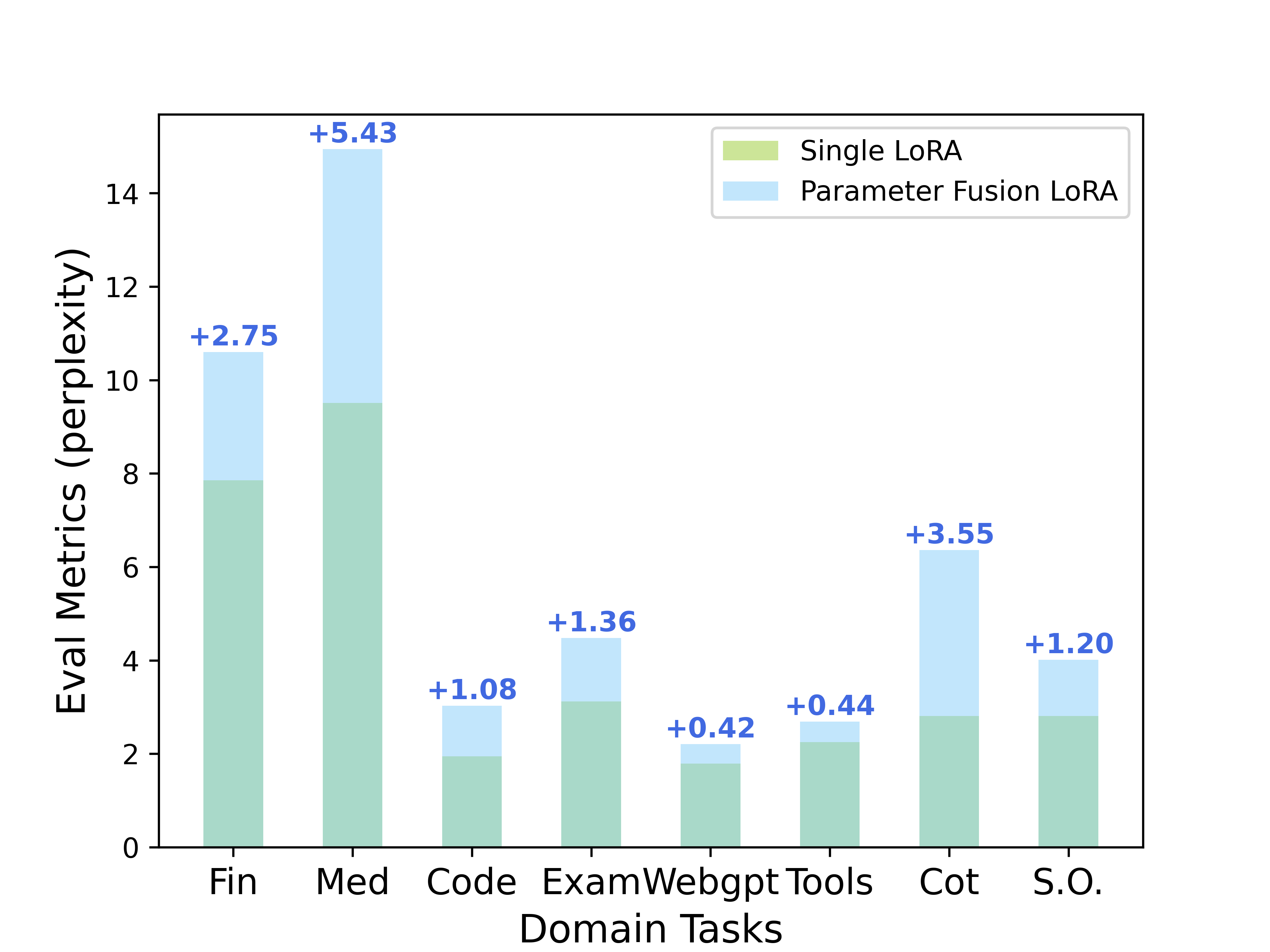}
\caption{Comparing different LoRA models using multi-task parameter fusion method and corresponding task method.} 
\label{fig:intro}
\end{figure}

Large language models (LLMs) have played a pivotal role in expediting the advancement of natural language processing (NLP), offering a versatile and task-agnostic foundation that underpins an extensive array of applications. The intrinsic diversity found in domain-specific data poses a substantial challenge in training a general-purpose base LLMs. Consequently, there has been a surge in the adoption of domain-specific LLMs to tackle intricate problems within specialized domains, such as SQL-PaLM~\citep{sun2023sqlpalm}, BloombergGPT~\citep{wu2023bloomberggpt}, ChatLaw~\citep{cui2023chatlaw}, pdfGPT~\citep{pdfgpt2023}. In real-world application scenarios, the demand often arises for a multitude of customized capabilities.
LLMs with multiple customized capabilities can efficiently address a diverse range of user problems, and each specific functional module can be optimized individually.

Domain specification techniques are key to make large language models disruptive in various applications~\citep{zhao2023domain}. To learn sufficient domain knowledge and not lose basic capability, adapter-based fine-tuning methods (e.g., Adapters~\citep{pmlr-v97-houlsby19a}, LoRA~\citep{hu2021lora}) introduce a limited number of domain-specific parameters to retain domain-related knowledge and do not need to fine-tuning all parameters of the pre-trained model, which can effectively reduce the training cost of LLMs.

In order to obtain multiple customized capabilities, the simplest and efficient fine-tuning approach is to directly mix data from multiple domains together and only add one LoRA module for instruction fine-tuning. The second is some two-stage approaches. 
We first train multiple domain LoRA modules individually, and then introduce a domain classifier to select appropriate LoRA model. In addition, \citet{pfeiffer2020adapterfusion,huang2023lorahub} add the \textit{AdapterFusion} layer and \textit{element-wise} LoRA composition to implicitly fuse parameter knowledge of multiple task adapters. The MoE~\citep{shen2023mixtureofexperts} introduces multiple experts to process different types of input data. \citet{llm-blender-2023} is to consider the outputs of ensemble LLMs comprehensively.

However, these approaches have two main issues. First, in the practical application scenario, there are often multiple destructive domain tasks with heterogeneous and imbalance training data, as well as limited computing resources. Therefore, the implicit parameter fusion methods exist mutual disturbance, which results in degraded model performance on in-domain tasks, as illustrated in Figure~\ref{fig:intro}. We often focus more on the domain-specific expertise of a domain-specific LLM than its generalization performance.
In addition, the MoE mechanism needs to be trained from scratch based on a new model structure and a large amount of training corpus. The ensemble LLMs require sufficient computing resources to deploy multiple independent LLMs simultaneously.

In this paper, we address these limitations and propose an end-to-end parameter-efficient tuning method designed for multi-task learning on LLMs, dubbed MoA. First, we design a routing mechanism within decoder-only model architecture to automatically select LoRA experts, which can be applied to current mainstream LLMs, and can simultaneously deploy LoRA modules of multiple tasks with limited computing resources (using the same LLM). Meanwhile, our comprehensive model can still achieve excellent performance on different types of tasks.
Additionally, to improve the efficiency of training and inference, we implement a parallel processing strategy of different domain samples within a batch during the training process, and a LoRA expert selection approach in the inference time.
Our approach leverages the power of different expert models and the base LLM, and the complementarity of knowledge in different domains.
In summary, our contributions are as follows:
\begin{itemize}
\item We propose a MoA architecture for efficient multitask fine-tuning, which can avoid the interference and data imbalance between heterogeneous tasks and easily perform iterative optimization of single task.

\item We implement an explicit routing strategy in the training process, which can leverage the knowledge complementarity to further improve the single task performance and ensures the inference efficiency.

\item Extensive experiments on various benchmarks are conducted to verify the effectiveness of our approach. Meanwhile, it is flexible to combine multiple domain-specific LoRAs to form a comprehensive LLM.
\end{itemize}

\section{Related Work}
\label{sec:relatedworsk}

\textbf{Domain Specialization of LLMs.} The approaches in LLM domain specialization can be categorized into three corresponding classes of approaches: \textit{external augmentation, prompt crafting, and model fine-tuning} \citep{zhao2023domain}. We focus on the third method, which involves updating the LLM’s parameters to incorporate domain-specific knowledge directly into the model. Because the current LLMs have billions of parameters and the phenomenon of catastrophic forgetting, we use adapter-based fine-tuning (e.g, Adapter, LoRA) to train multiple domain experts in advance on different task data. LoRA is a parameter-efficient fine-tuning method, which facilitates the adaptation of LLMs using a small-scale external module. As such, LoRA tuning presents a resource-efficient technique to quickly adapt LLMs for novel tasks with restricted training data. 

\noindent\textbf{Mixture-of-Experts.} The Mixture of Experts (MoE) is an ensemble method, often visualized as a collection of sub-modules, or ’experts’, each specializing in processing different types of input data. Each expert is controlled by a router that is then selectively activated based on the type of input data. This technique achieves excellent performance in other domains, including computer vision, speech recognition and multi-modal applications~\citep{fedus2022review}. Inspired by the idea of MoE, we regard each task's LoRA module as a domain expert. Meanwhile, we introduce a routing algorithm to make different domain data automatically choose respective expert, and hence experts in different domains can be combined into a comprehensive model. Furthermore, \citet{shen2023mixtureofexperts} also demonstrates that the combination of MoE and instruction tuning can improve task-specific performance.

\noindent\textbf{Multi-Task Composition.} 
The methods of two-stage learning or end-to-end multi-task learning are commonly used to obtain the combination of multi-task capabilities.
The two-stage method requires maintaining a specialized routing model to serve multiple task models, whose overall performance is limited to a single LoRA model. The end-to-end methods introduce new parameter layers or perform implicit parameter fusion. Specifically, \citet{pfeiffer2020adapterfusion} trains a fusion parameter layer to compose the information stored in the multiple task adapters. \citet{huang2023lorahub} uses a set of learnable weights to integrate multiple LoRA modules into a unified module. Although they can obtain some generalization ability through the combination of parameters trained on various tasks, these fusion methods result in performance degradation on the original task. \citet{zhao2023domain} also shows that it is difficult to effectively learn all specialized knowledge of different domains in one LLM.

\section{Methodology}
\begin{figure*}[ht]
\centering
\includegraphics[width=0.7\textwidth]{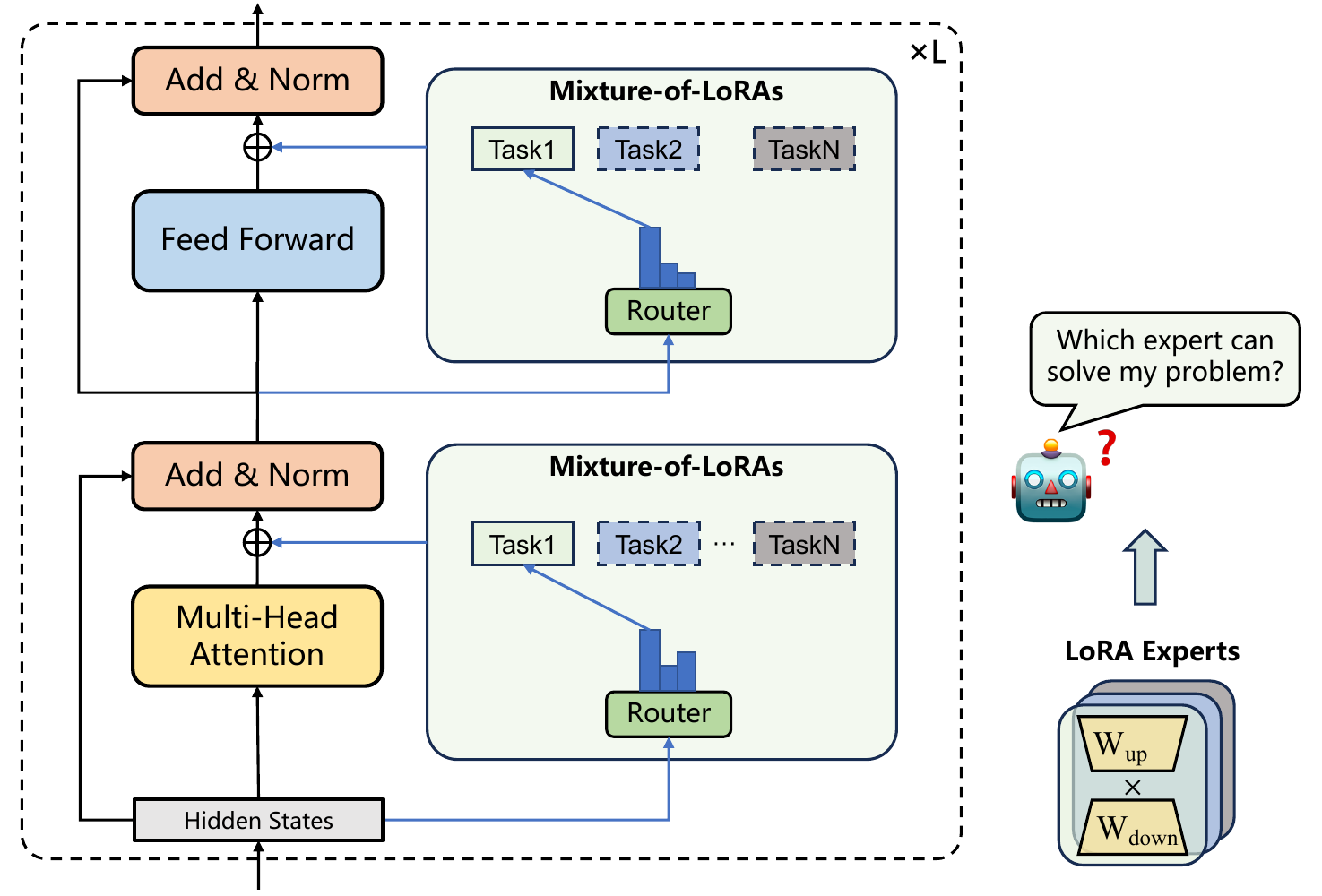}
\caption{Overall architecture of our proposed MoA.}
\label{fig:MoA}
\end{figure*}

The proposed Mixture-of-LoRAs (MoA) can be observed in Figure~\ref{fig:MoA}. Given the variations in data scales and training complexities, we commence by separately training \textit{N} LoRA modules $\{\mathcal{M}_1,...,\mathcal{M}_N\}$ for \textit{N} distinct task types $\{\mathcal{T}_1,...,\mathcal{T}_N\}$. It is worth noting that the task types mentioned here and the domain-specific data or tasks referred to in the paper have the same meaning, all representing scenario tasks that require certain expertise to solve. Initially, we obtain the optimal LoRA module parameters on each scenario task data. 
These modules demonstrate commendable performance within their respective domains. Subsequently, a routing mechanism is employed to integrate the \textit{N} LoRA modules under a shared LLM. Specifically, multiple LoRA modules are simultaneously incorporated alongside each transformer layer. Preceding each transformer layer, a routing mechanism is inserted to guide the selection of distinct LoRA experts.

\subsection{Learning algorithm}
In the first stage, we train a LoRA module for each of the \textit{N} tasks, which mitigates the problem of catastrophic forgetting of base LLM. Different tasks often have different data scales and training difficulties. Hence, each LoRA module needs to be optimized individually. These LoRA modules can be arbitrarily combined, added, or removed after initial training. This adapter schemes could enable more fine-grained control over which parts of the LLM are domain-specific.

In the second stage, the \textit{N} LoRA modules can be aligned with the MoE’s expert design. We combine the set of \textit{N} LoRAs using a routing strategy. While keeping the base LLM parameters $\Theta$ fixed, we introduce router parameters $\mathcal{R}$ that learn to select the appropriate expert for users’ target tasks (as shown in Figure~\ref{fig:MoA}). The trainable router and LoRA parameters are combined to jointly optimize the autoregressive language modeling tasks. The training data for this stage is evenly sampled from the original data of each task, obviating the need to acquire new, high-quality supervised data. The final loss of the MoA is the summation of the language modeling loss and the MoE routing loss:

\begin{equation}
   \mathcal{L}=\mathcal{L}_{LM}(\mathbf{x})+\eta\mathcal{L}_{cls}
\end{equation}
Here, $\eta$ is a parameter that controls the weight of the routing loss and the convergence rate, $\mathcal{L}_{cls}$ is the cross-entropy loss of expert classification, and $\mathcal{L}_{LM}(\mathbf{x})$ is defined as follows:
\begin{equation}
   \mathcal{L}_{LM}(\mathbf{x}) = \sum_{i=1}^n{logP(x_i|x_{<i})}
\end{equation}
In this context, the language modeling task (LM) is the commonly used objective for pretraining decoder-only LLMs (e.g., GPT-3~\citep{brown2020language}). Here, $\mathbf{x}$ represents a sequence of tokens $\{x_1,...,x_n\}$. The LM task involves predicting the target tokens $x_i$ autoregressively, based on the preceding tokens $x_{<i}$ within the sequence.

By dividing the process into two stages: (1) training domain experts in LoRAs, and (2) combining diverse capabilities through a routing strategy, we effectively address concerns such as catastrophic forgetting, task interference, and instability during multi-task training.

\subsection{Routing Strategy}
Prior approaches~\citep{fedus2022switch,lepikhin2020gshard} on routing strategy have typically focused on learning token-level weighting functions, often assigning one or two experts per token. This approach necessitates careful load balancing to ensure utilization of all experts, prompting the exploration of explicit balancing mechanisms~\citep{lewis2021base}.

In contrast, we adopt a sequence-level routing strategy that leverages domain metadata to route data to LoRA experts. While all token sequences traverse the weight matrices in LLM’s transformer layers, during training, each transformer layer employs a distinct router to assign training data labeled with its corresponding domain. The final routing loss enforces precise data-to-expert assignments for each data type. After training with a modest quantity of balanced data, a robust router is obtained. Meanwhile, multi-task training enhances the generalization of the original tasks.

Throughout the training process, language modeling and router classification tasks complement each other. However, we enhance expert selection during inference by employing techniques such as voting or selecting the last expert (as detailed in ~\ref{inference}). This optimization aims to enhance generation efficiency and contextual consistency.

\subsection{MoA Architecture}
We design an LoRA expert explicitly for each domain (i.e., eight experts for eight training domains in our multi-domain corpus). LoRA updates weights using the formula: 
\begin{equation}
	\begin{split}
	W_0+\mathbf{router}(W_r,W_1,...,W_N) \\
	= W_0+\mathbf{router}(W_r,A_1B_1,...,A_NB_N)
	\end{split}
\end{equation}
where $W_0 \in \mathbb{R}^{d\times k}$, $W_r \in \mathbb{R}^{hidden\_dim \times N}$, $A \in \mathbb{R}^{d\times r}$ and $B \in \mathbb{R}^{r\times k}$.
$W_0$ denotes the attention and feed-forward weight matrices of the base LLM, whose parameters are fixed. The 
parameter of router(·) is trainable. $W_r$ is the parameter of the router, which is implemented by a linear layer. The $A_iB_i$ defines the LoRA module $\mathcal{M}_i$, which is repeated multiple times within each transformer layer to reduce trainable parameters for adapting to different domain tasks. The multiple $\mathcal{M}_i$ experts are placed in parallel alongside $W_0$, departing from previous methods~\citep{pfeiffer2020adapterfusion,gururangan-etal-2022-demix,huang2023lorahub} that add shared fusion layers or replace dense feedforward layers with multiple experts’ feedforward networks. The routing algorithm is a key feature in all sparse expert architectures. We adopt an intuitive method to assign different experts to handle tasks in different domains, which helps avoid interference between tasks. Each transformer layer adds a router to select the most appropriate expert. Each router is implemented as a two-layer MLP. This implementation is simple and doesn’t significantly increase the number of training parameters.

\begin{table*}[htbp]
\footnotesize
\centering
\setlength{\tabcolsep}{4pt} 
\scalebox{0.95} {
\begin{tabular}{lcccccc}
\toprule
\textbf{Domain} & \textbf{Source} & \textbf{Language} & \textbf{\# Train (Eval.) Tokens} \\
\midrule\midrule
\textsc{Finance} & Financial related instructions~\citep{alpaca-cot} & EN & 1.2M (0.24M)  \\ 
\textsc{Medicine} & 10k real conversations between patients and doctors~\citep{li2023chatdoctor}& EN & 1.4M (0.28M)  \\ 
\textsc{Leetcode} & \multirow{2}{*}{Chinese Open Instruction Generalist~\citep{zhang2023chinese}}  & CN & 9.3M (2.09M)  \\
\textsc{Exam} &  & CN & 3.6M (0.71M) \\
\textsc{webgpt} & Retrieval question answering dataset~\citep{nakano2021webgpt} & EN & 7.4M (1.46M)  \\
\textsc{Gpt4tools} & A collection of tool-related instructions~\citep{gpt4tools} & EN & 7.5M (1.49M)  \\
\textsc{Cot} & Several Chain-of-Thought datasets~\citep{longpre2023flan} & EN & 1.1M (0.22M)   \\ 
\textsc{Stackoverflow} & 57k dialogs from StackOverFlow questions~\citep{xu2023baize} & EN & 0.9M (0.18M)   \\ 
\bottomrule
\end{tabular}
}
\caption{Statistics of SFT datasets. \# Train (Eval.) Tokens denotes the size of our training and evaluation (i.e. validation and test) data, obtained via the Qwen tokenizer. We evenly sample about 10k training data and 2k validation data on each dataset.} 
\label{tab:datasets}
\end{table*}

\section{Experiments}

\subsection{Experimental Setup}
\textbf{Datasets.} To evaluate the effectiveness of MoA, we first conduct experiments on various supervised fine-tuning (SFT) datasets of heterogeneous domains. \textsc{Finance},  \textsc{Medicine} and \textsc{Leetcode} belong to the specialized domain dataset. \textsc{Exam}, \textsc{Webgpt} and \textsc{Gpt4tools} limit the output format of the LLM and allow the model to learn special functions. Other datasets include chain-of-thought, dialog, etc. Meanwhile, both English and Chinese are involved. We show the statistics of all datasets in Table~\ref{tab:datasets}.

\noindent\textbf{Implementation Details.} We use the \textit{Qwen-7b} as our base LLM, which is the decoder-only architecture. We use the vocabulary of 151,851 BPE types, and train with 4,096-token sequences. We set the total number of training steps based on this allocated runtime, set 10\% of these steps to be warmup, and use the AdamW~\citep{loshchilov2017decoupled} optimizer with a cosine learning rate decay. Learning rate is set to $1e^{-5}$. Each worker processes eight sequences of length 4,096, and gradients are accumulated over 4 updates. We clip gradients if their L2 norm exceeds 0.1. In the inference time, we report test perplexity after a single run of training on 8 NVIDIA A100 80GB GPUs.

\subsection{Models and Metrics }
\textbf{Single-\textsc{LoRA}} The first baseline is a LoRA trained on data within the domain, which could be viewed as a specialized model for handling domain tasks.

\noindent\textbf{Single-\textsc{LoRA}} \textbf{(mixed)} We train a single LoRA on mixed data from all domains. While there is no explicit conditioning on domain, the gradient updates during training average across all domains represented in a batch.

\noindent\textbf{\textsc{MoA}} We add multiple domain LoRAs in the transformer based on the first baselines, and use routing strategy and domain label information for multitask training. Under the \textit{routing} setting, the test data domain is unknown.

\noindent\textbf{MoE-\textsc{LoRA}} We leverage the idea of sparsely activated Mixture-of-Experts (MoE). We add multiple domain LoRA modules on the bypass of transformer layer. Each input token will be processed by a limited subset of experts. Different from our MoA, this approach can add any number of experts and has no concept of data domain.

\noindent\textbf{MoE-\textsc{LoRA}} \textbf{(naive)} The architecture of this model is exactly the same as MoE-LoRA, and the only difference is that we randomly initialize all LoRA modules.

\noindent To comprehensively evaluate the performance of different models, we use several evaluation metrics, including the perplexity (PPL) of generated texts, and the bilingual evaluation understudy (BLUE) and the longest common subsequence (ROUGE-L) between the generated answer and the gold answer.

\subsection{Main results}\label{main_results}

\begin{figure}[ht]
\centering
\includegraphics[width=0.47\textwidth]{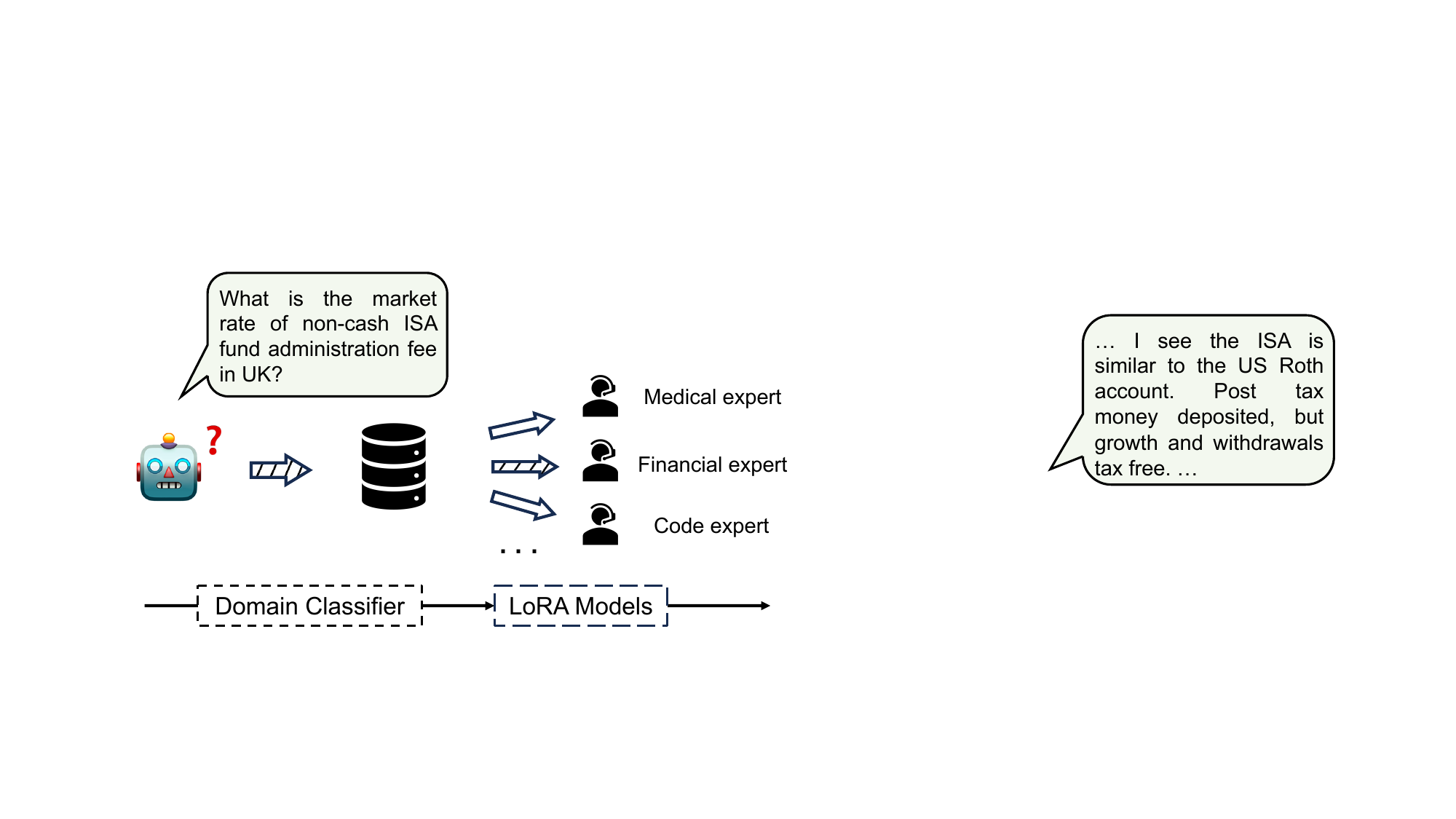}
\caption{The flow diagram of classifier+LoRAs.}
\label{fig:Two-stage}
\end{figure}

\textbf{Classifier+LoRAs} The most intuitive method of integrating multiple LoRA experts is to use a specific classifier to act as a distributor, as shown in Figure~\ref{fig:Two-stage}. We train a classifier based on roberta-base~\citep{liu2020roberta} using the same training data as MoA. The specific classification performance is shown in Table~\ref{tab:classification}. This approach is so flexible that we can combine multiple LoRA modules and avoid interference between tasks. However, the performance of this approach is limited to each LoRA expert. From the classification results, our router performs better than the classifier overall.

\begin{table*}[htbp]
 \small
 \centering
\begin{tabular} {@{}cccccccccc @{}} 
\toprule
\multirow{2}{*}{Domain}  & \multicolumn{3}{c}{Single-LoRA}  & \multicolumn{3}{c}{Single-LoRA (mixed)} & \multicolumn{3}{c}{MoA}\\ 
\cmidrule(l){2-10}  & \multicolumn{1}{c}{PPL} & \multicolumn{1}{c}{BLUE} & \multicolumn{1}{c}{ROUGE-L} & \multicolumn{1}{c}{PPL} & \multicolumn{1}{c}{BLUE} &\multicolumn{1}{c}{ROUGE-L} & \multicolumn{1}{c}{PPL} & \multicolumn{1}{c}{BLUE} &\multicolumn{1}{c}{ROUGE-L}\\ 
\midrule
\multicolumn{1}{l}{\textsc{Finance}} & 7.8479 & 18.5975 & 28.6266 & 7.7214 &\textbf{22.4846} & \textbf{32.5574} & \textbf{7.5287} &20.5774 &30.6797\\
\multicolumn{1}{l}{\textsc{Medicine}} & 9.5097 & 13.6096 & 18.8911 &9.0499 &13.5373 &19.4425&\textbf{8.4561}& \textbf{13.8811}&\textbf{19.8118} \\
\multicolumn{1}{l}{\textsc{Leetcode}} & 1.9527 & 34.8582& 47.8152 &2.0289 & 35.2886 & 46.6290 & \textbf{1.9311}&\textbf{37.4872}&\textbf{49.3256}\\
\multicolumn{1}{l}{\textsc{Exam}} & 3.1154 &3.0871 &18.5609&3.1135 & 4.3259& 16.6206& \textbf{2.9752}&\textbf{4.7942}&\textbf{19.1840} \\
\multicolumn{1}{l}{\textsc{Webgpt}} & 1.7945 &38.8995& 41.4447& 1.8484 &39.6297& 42.0700 & \textbf{1.7933}&\textbf{40.2602}& \textbf{43.7395}\\
\multicolumn{1}{l}{\textsc{Gpt4tools}} & 2.2525& 64.7501& 71.4391& 2.2497& 66.3450& 73.1289& \textbf{2.2123}&\textbf{69.2596}& \textbf{74.5962}\\
\multicolumn{1}{l}{\textsc{Cot}} & 2.8126&34.5210&45.7961 & 2.6474 &43.6290&\textbf{53.2125}&\textbf{ 2.5931}&\textbf{40.2529}&50.3844 \\
\multicolumn{1}{l}{\textsc{S.O.}} & \textbf{2.8169} &19.9554& 29.7282& 2.9012 & 19.4896 &28.4694 & 2.8999& \textbf{23.0412}& \textbf{31.9793}\\
\midrule
\multicolumn{1}{l}{\textbf{Average}} & 4.0128 &28.5348 &37.7877&3.9450 &30.5912& 39.0163 & \textbf{3.7987} & \textbf{31.1942}& \textbf{39.9626}
\\\bottomrule
\end{tabular}
 \caption{In-domain test-set performance for different training strategies of LoRA. We report PPL, tokenized BLUE and ROUGE-L above. Both Single-LoRA (mixed) and MoA are tested under the test data of unknown domain labels and Single-LoRA is tested on the test data of the corresponding domain. PPL is short for perplexity. The best value per metric on different tasks is in bold.}
 \label{tab:main-result}
\end{table*}

\begin{table}[htbp]
 \small
 \centering
\begin{tabular} {@{}cccc @{}} 
\toprule
\multicolumn{1}{l}{\textbf{Domain}} & \textbf{test size}  & \textbf{Classifier} & \makecell[c]{\textbf{Router}} \\
\midrule
\multicolumn{1}{l}{\textsc{Finance}} & 2000 & 98.80\% &99.60\% \\
\multicolumn{1}{l}{\textsc{Medicine}} & 1221  &99.92\% & 99.92\% \\
\multicolumn{1}{l}{\textsc{Leetcode}} & 1952 &99.95\% & 100.00\% \\
\multicolumn{1}{l}{\textsc{Exam}} & 1999 &99.95\%&  100.00\% \\
\multicolumn{1}{l}{\textsc{Webgpt}} & 2000 &100.00\%  & 99.85\% \\
\multicolumn{1}{l}{\textsc{Gpt4tools}} & 2000 & 100.00\% &100.00\% \\
\multicolumn{1}{l}{\textsc{Cot}} & 2000   &99.75\% & 99.95\%\\
\multicolumn{1}{l}{\textsc{Stackoverflow}} & 2000   &99.00\%& 99.90\%\\
\midrule
\multicolumn{1}{l}{\textbf{Average}} & 1896.5   & 99.67\%& \textbf{99.90\%}
\\\bottomrule
\end{tabular}
 \caption{The classification accuracy of MoA router and a specific classifier by domain at inference time.}
 \label{tab:classification}
\end{table}

\noindent\textbf{End-to-end methods} The end-to-end methods mean that one model can directly solve multiple tasks even if the test data domain is unknown. Table~\ref{tab:main-result} shows that test perplexities, blue-4 and rouge-l, averaged across the eight training domains. Training in the mixed domain data is helpful for the overall performance (Perplexity: 4.0128$\rightarrow$3.9450, BLUE: 28.5348$\rightarrow$30.5912, ROUGE-L: 37.7877$\rightarrow$39.0163). However, the performance decreases on data with strict output formats such as \textsc{Webgpt}, \textsc{Stackoverflow}, which also shows that not all additional domain information is complementary. We hypothesize that separate training is advantageous for heterogeneous domains. Therefore, we design an efficient multi-task learning method to avoid interference between partial tasks. Our approach achieves significant improvements across all datasets.

\begin{table}[htbp]
 \small
 \centering
\begin{tabular} {@{}ccccc @{}} 
\toprule
\multicolumn{1}{c}{Model} & \makecell[c]{LoRA} & \makecell[c]{LoRA\\(mixed)} & \makecell[c]{MoA}  \\
\midrule
\multicolumn{1}{l}{\makecell[c]{\textbf{trainable}\\\textbf{parameters}}} & 143M & 143M & 143M*8+1.05M\\
\bottomrule
\end{tabular}
 \caption{The trainable parameters under different LoRA combinations. The router module takes up only 1.05M parameters.}
 \label{tab:parameter}
\end{table}

In order to further validate the reliability of the MoA's performance, we conducts accuracy evaluation experiments on datasets with standard answers. The \textsc{Exam} dataset is utilized for the experiments, as it primarily consists of secondary and university entrance exam questions, with the majority being multiple-choice questions (including single and multiple selections) and a small portion of true/false questions. Accuracy is calculated by directly comparing the answers with the reference answers. Specifically, string processing functions, regular expressions, etc., are used to parse out specific options (A/B/C/D) or judgment results (T/F) from the answers, followed by calculating the accuracy of responses. The final results are presented in Table~\ref{tab:exam_acc}. Despite the overall low accuracy due to the difficulty of the questions, the accuracy of MoA is significantly higher than the other two models (\textbf{+5.86}\%, \textbf{+5.48}\%).
\begin{table}[htbp]
 \small
 \centering
\begin{tabular} {@{}cccc @{}} 
\toprule
\multicolumn{1}{l}{Model} & Total  & Right & Accuracy \\
\midrule
\multicolumn{1}{l}{Single-LoRA (mixed)} & 1331  & 515 & 38.69\% \\
\multicolumn{1}{l}{Single-LoRA} & 1331 & 520 & 39.07\% \\
\multicolumn{1}{l}{MoA} & 1331 & 593 &\textbf{44.55\%}
\\\bottomrule
\end{tabular}
\caption{The accuracy of responses on the Exam test dataset.}
 \label{tab:exam_acc}
\end{table}

In addition, this subsection also introduces GPT-4 as an evaluation expert to assess \textsc{Finance}, \textsc{Medicine}, and \textsc{Webgpt} datasets. As the answers in these datasets do not follow a fixed pattern, this subsection adopts the common evaluation method in the community of LLMs. Evaluation scoring is conducted through a larger model. In this study, GPT-4 is utilized to provide accuracy scores for the model's responses to questions and standard answers. The complete evaluation prompt is illustrated in Appendix~\ref{sec:appendixA}.

Apart from the three models: Single-LoRA, Single-LoRA (mixed), and MoA, we also evaluates the individual LoRA modules within the MoA model. These are represented in the table~\ref{tab:moa_score} as Single-LoRA of MoA. 
Due to fluctuations in scoring by the large language model, each scoring process is invoked three times and the average is taken. From the experimental results, it is observed that after multi-task learning training within MoA, the performance of each LoRA module surpasses the original Single-LoRA modules in each task. When handling specific professional tasks individually, these modules exhibit outstanding performance. Therefore, through multi-task learning methods, the performance of multiple LoRA modules is further enhanced while effectively collaborating.
\begin{table}[htbp]
 \small
 \centering
\begin{tabular} {@{}cccc @{}} 
\toprule
\diagbox{Model}{Score}{Dataset} & Finance  & Medicine & Webgpt \\
\midrule
\multicolumn{1}{l}{Single-LoRA (mixed)} & \textbf{76.91}  & 57.49 & 87.92 \\
\multicolumn{1}{l}{Single-LoRA} & 75.99 & 57.11 & 88.59 \\
\multicolumn{1}{l}{Single-LoRA of MoA} & 76.30 & \underline{58.01} & \underline{89.00} \\
\multicolumn{1}{l}{MoA} &  \underline{76.56} & \textbf{60.68} &\textbf{89.27}
\\\bottomrule
\end{tabular}
\caption{The evaluation scoring of the GPT-4 on the Finance, Medicine, and WebGPT datasets. The highest value per column is in bold and the second highest value is \underline{underlined}.}
 \label{tab:moa_score}
\end{table}

Based on the metrics such as PPL, BLUE, ROUGE, and accuracy metrics for specific downstream tasks presented in the paper, we can conclude that our proposed simple and efficient architecture effectively enables learning of various domain-specific competencies within a single large language model, while also avoiding interference between different tasks. Furthermore, each functional module is relatively independent, facilitating efficient consolidation of additional data for further optimization. Additionally, this method significantly saves computational resources during deployment.

\subsection{Mixing LoRA Experts at Inference Time}\label{inference}
The previous section establishes that our multi-task training method improves the performance of the single LoRA expert on test data. In addition, the mixing of multi-domain data is effective in the training process. In practice, however, the original training data scale of the respective tasks is relatively large and textual data to be evaluated may not come with a domain label.

In these cases, the mixed training of a large amount of data is unfavorable for our later iterations of single tasks. We propose to treat $LoRA_1,...,LoRA_N$ as domain experts, transforming the input text into a matter of expert selection. The routing strategy is introducing to solve the problem of interference between tasks. The parameter of router is shown in Table~\ref{tab:parameter}, which only accounts for a very small percentage. Our approach is parameter-efficient and selective complementarity between tasks. The results of evaluation on the test data of unknown domain are shown in Table~\ref{tab:main-result}. 


\begin{figure*}[ht]
\centering
\includegraphics[width=0.90\textwidth]{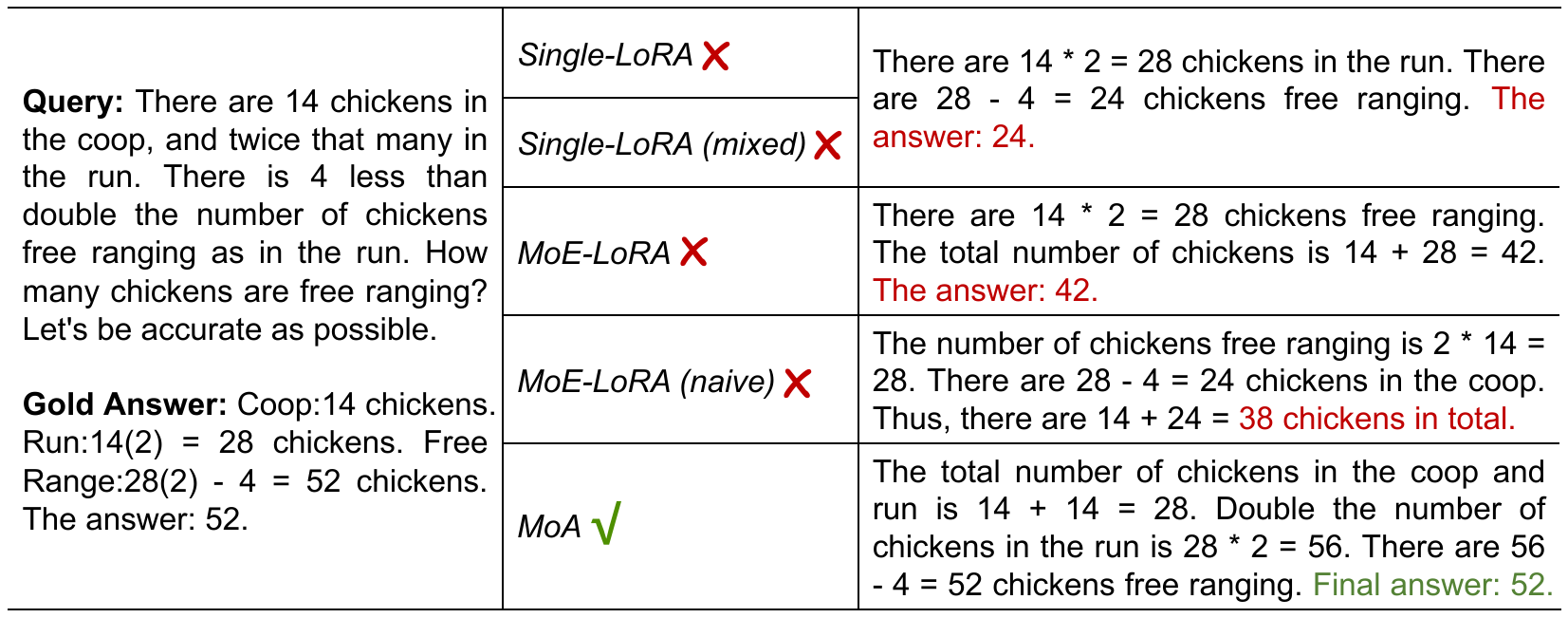}
\caption{Case study of the predicted output of different models.}
\label{fig:case-study}
\end{figure*}

To further explore the actual effect of the routing module, we test its selection ability in the inference process. The number of our routers is consistent with the number of transformer layers in our LLM. Each router is implemented with a linear layer. Therefore, to improve the efficiency of the inference process and the consistency of generation, we choose to use (1) voting of all routers and (2) the result of last router as the selected LoRA module in the subsequent inference process. The experimental results show that the classification performance of the last router is more stable. Finally, we only use 1.05M routing parameters to achieve an average accuracy of 99.90\% in Table~\ref{tab:classification}. Overall, the router plays two key roles in our model architecture. One is to learn the complementarity of non-heterogeneous domain knowledge and in the training process. The other is to select appropriate LoRA expert to solve problems in unknown domains in the inference process.

\begin{table}[htbp]
\footnotesize
\centering
\setlength{\tabcolsep}{4pt} 
\scalebox{1.05} {
\begin{tabular}{lccc}
\toprule
Methods & PPL & BLUE & ROUGE-L  \\
\midrule
MoE-LoRA & 3.8578  & 29.1640 & 37.5960\\
MoE-LoRA (naive)  & \textbf{3.7969 } & 29.4170 & 37.3917  \\
\midrule
MoA & 3.7987 &\textbf{ 31.1942} & \textbf{39.9626}\\
\bottomrule
\end{tabular}}
\caption{The averaged test performance comparison on eight tasks. All of the above methods have the same number of LoRA experts and have exactly the same model parameters.}
\label{tab:ablation}
\end{table}

\subsection{Ablation Studies}\label{ablation}
We further conduct specific experiments to investigate the effectiveness of different components in our model.

\noindent\textbf{Impact of Domain Label Information} In addition to the language modeling loss, we also add the domain label routing loss in the training process. To verify the effectiveness of our proposed mechanism, we remove the routing loss and our model devolved into MoE-LoRA. Therefore, the MoE-LoRA does not introduce explicit domain label information in the training and inference process and guarantees the same number of parameters as MoA. As shown in Table~\ref{tab:ablation}, MoA has achieved an overall improvement over MoE-LoRA, which demonstrates that the domain label information is useful for different tasks. Furthermore, we evaluate the performance of single LoRA module after multi-LoRA joint training in Table~\ref{tab:single-lora}. The MoA further improves the perplexity performance of each LoRA module from 4.0128 to 3.7962, while MoE-LoRA causes a slight decrease. Therefore, 
when we need to expand to more tasks or combine multiple functions, MoA is more flexible and effective.

\begin{table}[htbp]
 \footnotesize
 \centering
\begin{tabular} {@{}cccc@{}} 
\toprule
Domain & \multicolumn{1}{c}{Single-LoRA} & \multicolumn{1}{c}{MoE-LoRA}& \multicolumn{1}{c}{MoA} \\ 
\midrule
\multicolumn{1}{l}{\textsc{Finance}} & 7.8479 & 7.6623 & \textbf{7.5235}\\
\multicolumn{1}{l}{\textsc{Medicine}} & 9.5097 & 9.6510 & \textbf{8.4488}\\
\multicolumn{1}{l}{\textsc{Leetcode}} & 1.9527 & 2.0087 &\textbf{1.9296}\\
\multicolumn{1}{l}{\textsc{Exam}} & 3.1154 & 3.1455 & \textbf{2.9745}\\
\multicolumn{1}{l}{\textsc{Webgpt}} & 1.7945 & 1.8080 & \textbf{1.7927}\\
\multicolumn{1}{l}{\textsc{Gpt4tools}} & 2.2525 & 2.2524 & \textbf{2.2123}\\
\multicolumn{1}{l}{\textsc{Cot}} & 2.8126 & 2.9205 & \textbf{2.5910}\\
\multicolumn{1}{l}{\textsc{S.O.}} & \textbf{2.8169} & 2.8801 & 2.8968\\
\midrule
\multicolumn{1}{l}{\textbf{Average}} & 4.0128 &4.0411 &\textbf{3.7962}\\
\bottomrule
\end{tabular}
 \caption{The test perplexity of corresponding LoRA module in different models on each task dataset.}
 \label{tab:single-lora}
\end{table}

\noindent\textbf{Impact of Different Initialization Methods} To evaluate the effectiveness of various LoRA initialization methods, we present the MoE-LoRA (naive) approach. In contrast to MoE-LoRA, the LoRA modules in the naive variant are initialized randomly. This approach leverages the complete dataset from all domains, leading to a significant increase in total training duration. However, this approach proves to be highly inefficient when we require customization of different capability combinations. Moreover, the comparative analysis presented in Table~\ref{tab:ablation} demonstrates that both methods yield comparable results across eight tasks. Therefore, conducting further training on the initial multi-domain LoRA parameters emerges as a highly efficient method.

\subsection{Case Study}
Figure~\ref{fig:case-study} shows the comparison of the outputs of different models on a specific reasoning question. 
From the above results, only the MoA correctly understands the multiple relationships in the query and predicts the right answer, while other approaches are slightly less capable of reasoning. Furthermore, the MoA is superior to the single-LoRA, which also illustrates the advantages of our multitask fine-tuning. This approach can not only avoid the interference between different tasks, but also further improve the performance on a single task, so it has a great application prospect.

\section{Conclusions}
\label{sec:conclusion}
We introduce MoA architecture, which provide an efficient multi-task fine-tuning method for LLM, addressing interference among tasks and training instabilities. Each LoRA model can be iterated individually to quickly adapt to new domains. Meanwhile, MoA uses routing strategy to flexibly select the appropriate LoRA expert to solve the problem. It can arbitrarily combine multiple domain-specific LoRA modules to implement a LLM with multiple specific capabilities. Future work may focus on how to flexibility add or remove LoRA modules with unsupervised learning, optimize the current routing algorithm, or reduce the scale of training data  in domain specialization of LLMs.



\bibliography{anthology,custom}

\newpage
\appendix
\section{Evaluation prompt}
\label{sec:appendixA}
The complete prompt used for invoking the GPT-4 API is as follows:

\noindent eval\_prompt = """ You will receive three parts of content: the questioner's question, the user's answer, and the reference answer.\\
Your task is to score the accuracy of the user's answer based on the following criteria.
Please ensure that you read and understand these instructions carefully.\\
Evaluation Criteria:\\
Accuracy - Whether the user's answer is consistent with the reference answer and has addressed the questioner's question. We define this dimension as 'whether the user's answer includes all the key points from the reference answer and has addressed the questioner's question.'\\
Evaluation Steps:\\
1. Carefully read the questioner's question, understand the key points of the question.\\
2. Carefully read the reference answer, understand the key points related to the question contained in the reference answer.\\
3. Check if the user's answer includes the key points from the reference answer and has addressed the questioner's question.\\
4. Based on the evaluation criteria, score within a range of 0 to 100, where 0 means the user's answer does not contain any key points from the reference answer and has completely failed to address the questioner's question; 100 means the user's answer includes all the key points from the reference answer and has correctly and completely addressed the questioner's question. \\
Example: Questioner's question: {{[query]}} User's answer: {{[llm\_answer]}} Reference answer: {{[fact]}} 
Evaluation result (score only): \\
Accuracy (0-100): """

where "[query]" represents user input, "[llm\_answer]" denotes the output of the MoA model, and "[fact]" stands for the standard answer provided by the dataset.

\end{document}